\documentclass[10pt,twocolumn,letterpaper]{article}

\usepackage{cvpr}              %

\newcommand{\Fig}[1]{Fig.~\ref{fig:#1}}

\newcommand{\Tbl}[1]{Tab.~\ref{tab:#1}}

\usepackage{url}
\usepackage{multirow}
\usepackage{graphicx}
\usepackage{stfloats, cuted, lipsum}
\usepackage{color, colortbl}
\usepackage{graphicx}
\usepackage{booktabs}
\usepackage{dsfont}
\usepackage{amsmath}
\usepackage{amssymb}
\usepackage{xfrac}
\usepackage{algorithm2e}

\definecolor{cvprblue}{rgb}{0.21,0.49,0.74}

\definecolor{gray}{rgb}{0.9, 0.9, 0.9}

\usepackage[pagebackref,breaklinks,colorlinks,allcolors=cvprblue]{hyperref}

\def\paperID{14081} %
\def\confName{CVPR}
\def\confYear{2025}

\title{Leveraging 3D Geometric Priors in 2D Rotation Symmetry Detection
}

\author{Ahyun Seo \quad \quad \quad \quad Minsu Cho\vspace{0.15cm}\\
Pohang University of Science and Technology (POSTECH), South Korea\\
{\small \url{http://cvlab.postech.ac.kr/research/RotSymDETR}}
}

\begin{document}
\maketitle
\begin{abstract}
Symmetry plays a vital role in understanding structural patterns, aiding object recognition and scene interpretation. This paper focuses on rotation symmetry, where objects remain unchanged when rotated around a central axis, requiring detection of rotation centers and supporting vertices. Traditional methods relied on hand-crafted feature matching, while recent segmentation models based on convolutional neural networks (CNNs) detect rotation centers but struggle with 3D geometric consistency due to viewpoint distortions. 
To overcome this, we propose a model that directly predicts rotation centers and vertices in 3D space and projects the results back to 2D while preserving structural integrity. By incorporating a vertex reconstruction stage enforcing 3D geometric priors—such as equal side lengths and interior angles—our model enhances robustness and accuracy. Experiments on the DENDI dataset show superior performance in rotation axis detection and validate the impact of 3D priors through ablation studies.
\end{abstract}

\section{Introduction}
\label{intro}

\begin{figure}[t]
    \centering
    \includegraphics[width=0.47\textwidth]{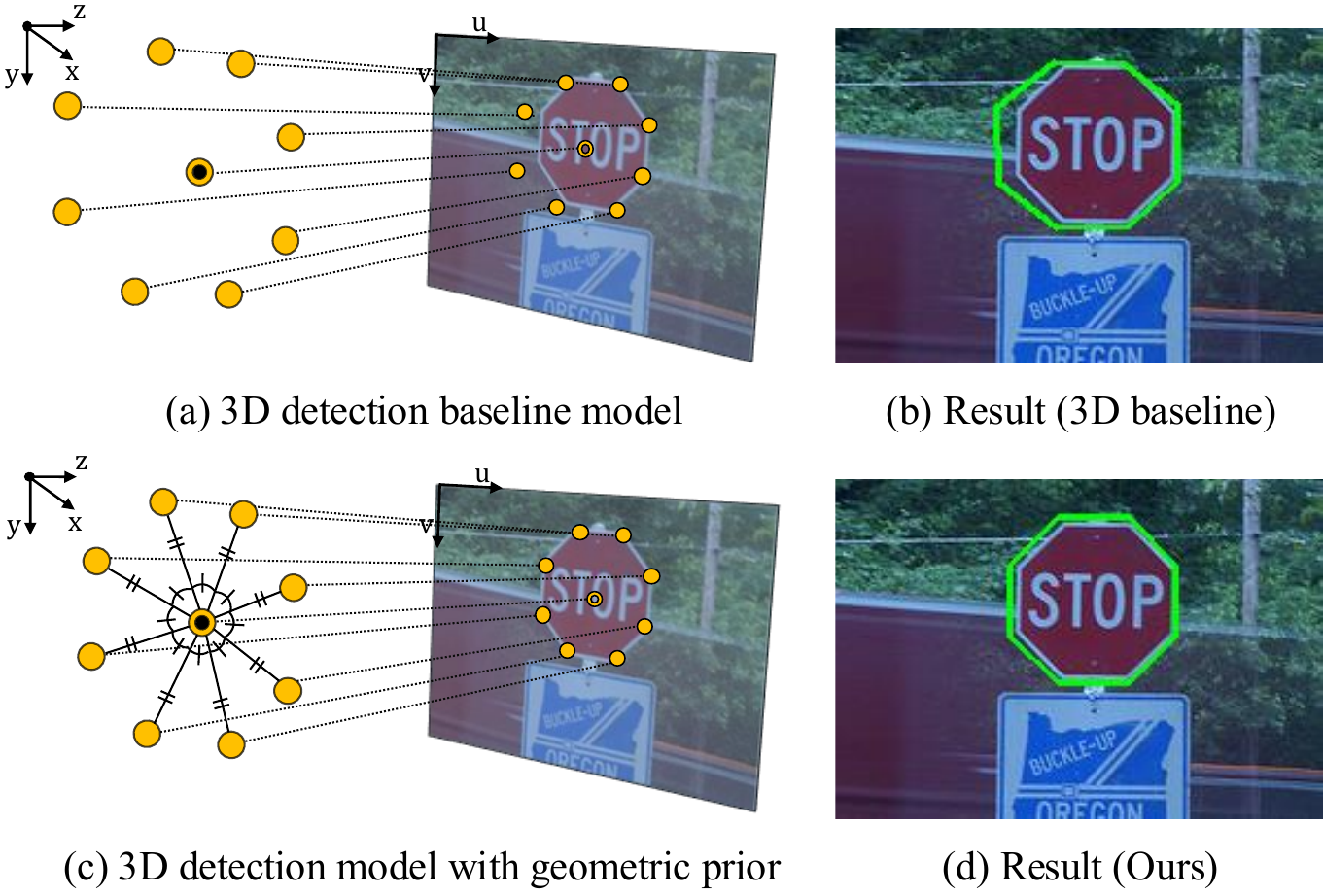}
   \caption{
    \textbf{Rotation symmetry detection models and results.}  
    (a) 3D detection baseline model without geometric priors, and (b) its qualitative results.  
    (c) Our 3D detection model with geometric priors, and (d) its corresponding qualitative results.  
    The results highlight the benefits of incorporating 3D geometric constraints.
}  \label{fig:teaser}
\vspace{-2mm}
\end{figure}

Symmetry is a key visual cue in both natural and man-made objects, helping us understand structure. Detecting symmetry aids tasks like object recognition and scene understanding by revealing consistent patterns across orientations. This paper focuses on rotational symmetry—when an object looks the same after rotation around a central axis. Detecting it involves predicting the rotation center and supporting vertices, providing a 3D interpretation of symmetry. 
The main challenge in this task lies in the annotation process: labels are assigned from a human perspective to capture symmetry in 3D, even when it is not apparent in 2D.

Traditional methods have explored symmetry characteristics such as rotation centers, symmetry group cardinality, and types~\cite{lee2008rotation, lee2009skewed}, relying on hand-crafted feature extraction and matching. More recently, convolutional neural networks (CNNs) have improved the robustness of rotation symmetry detection on real-world datasets~\cite{seo2022equisym, funk2017beyond}. However, these methods focus mainly on detecting rotation centers while overlooking supporting vertices and symmetry groups. Additionally, their reliance on image segmentation limits their ability to analyze individual symmetries.

A key limitation of 2D-based rotation symmetry detection models is their inability to enforce inherent geometric properties, as real-world objects are often labeled from a 3D perspective. This discrepancy leads to 2D ground truths that fail to capture geometric constraints due to viewpoint variation, requiring the model to infer 3D semantics from image features alone. While 2D models can be adapted for detection of individual axes, enforcing 3D geometric constraints under perspective distortions remains challenging.

To overcome this, we propose a 3D-based detection framework that directly predicts rotation centers and vertices in 3D space before projecting them back to 2D. Different approaches to rotation symmetry detection are illustrated in~\Fig{teaser}. (a) presents our 3D detection baseline, which improves upon 2D formulations but lacks explicit geometric constraints. Our final model (c) integrates 3D geometric priors, ensuring structural consistency and greater robustness to viewpoint variations. Qualitative comparisons in (b) and (d) demonstrate how these priors enhance the accuracy of rotation symmetry detection across perspectives.

To this end, we introduce a novel rotation symmetry detection model that leverages 3D geometric priors for a more comprehensive analysis of symmetry. Beyond detecting the rotation center, the model predicts rotation folds and supporting vertices, offering a richer representation of symmetry. Instead of regressing 2D coordinates, it outputs 3D coordinates for the rotation center and vertices, ensuring that symmetry is preserved in 3D space.
The key component of our model is the vertex reconstruction stage, which predicts a seed point and rotation axis to generate vertices while enforcing geometric constraints, eliminating the need for individual vertex predictions. Our model outperforms the state-of-the-art in rotation symmetry detection on the heatmap-based DENDI benchmark, and ablations confirm the effectiveness of its components for rotation vertex detection.

Our contributions are summarized as follows:
\begin{itemize}
  \item We introduce a novel rotation symmetry detection model that predicts sets of rotation centers, folds, and supporting vertices within a detection framework.
  \item We design a vertex reconstruction module that generates all vertices from a seed point and rotation axis, enforcing 3D geometric priors without per-vertex regression.
  \item Our approach achieves state-of-the-art results on the DENDI benchmark, with ablation studies validating the effectiveness of each proposed component.
\end{itemize}

\section{Related Work}
\label{related}
\paragraph{Rotation symmetry detection.}
Rotation symmetry is often detected as periodic signals using spatial autocorrelation~\cite{lin1997extracting, liu2004computational} or frequency-based methods such as spectral density and angular correlation~\cite{lee2008rotation, lee2009skewed, keller2006signal}. Loy and Eklundh~\cite{loy2006detecting} leverage SIFT descriptors, normalized by dominant orientation, to match keypoints and identify rotational symmetries. Other methods utilize Gradient Vector Flow (GVF)\cite{prasad2005detecting}, affine-invariant contour features\cite{wang2015reflection}, and polar domain representations~\cite{akbar2023detecting}.
Recent advances~\cite{seo2022equisym, funk2017beyond} have applied deep CNNs for rotation center detection to tackle real-world challenges. More recently, research has shifted toward 3D rotational symmetry detection~\cite{shi2022learning, sawada2022visual, hruda2022rotational}, given the complexities of 2D-based approaches. Motivated by the 3D-informed labeling of real-world 2D symmetry datasets, we propose a rotation symmetry detection model with 3D geometric priors for improved accuracy and robustness.

\paragraph{Set-based object detection.}
DETR~\cite{carion2020detr} reformulates object detection as a set-matching task, leveraging a Transformer~\cite{vaswani2017attention} to model interactions between features and objects. By directly assigning predictions to ground-truth boxes, DETR removes the need for post-processing to eliminate duplicate detections. To improve training efficiency, Deformable DETR~\cite{zhudeformable} introduces a deformable attention mechanism that focuses on relevant features. SOLQ~\cite{dong2021solq} further extends DETR’s approach by using object queries to simultaneously perform classification, bounding box regression, and instance segmentation. Other advancements in this framework include conditional spatial queries~\cite{meng2021conditional}, dynamic attention~\cite{dai2021dynamic}, unsupervised pre-training~\cite{dai2021up}, and query denoising~\cite{li2022dn}. In this paper, we design a set-based framework for rotation symmetry detection.

\paragraph{Vision-based 3D perception.}
Due to the high cost of LiDAR sensors, research has increasingly focused on camera-only 3D perception, projecting 2D image features into 3D space. LSS~\cite{philion2020lift} introduced a multi-view BEV approach with non-parametric depth, later improved for scalability~\cite{xie2022m}, geometry awareness~\cite{yang2023parametric}, and depth-supervised 3D object detection~\cite{huang2021bevdet, li2023bevdepth}.
DETR3D~\cite{wang2022detr3d} enhances BEV object detection with 3D-to-2D queries, while PETR~\cite{liu2022petr} encodes 3D positional information into 2D features, reducing projection errors. BEVFormer~\cite{li2022bevformer} improves view transformations with spatial cross-attention and temporal self-attention for multi-frame fusion. Inspired by~\cite{li2022bevformer}, this work introduces camera-centric queries to enable 3D rotation symmetry detection from 2D features.

\section{Proposed Method}
\label{method}
\begin{figure*}[t]
    \centering
    \includegraphics[width=\textwidth]{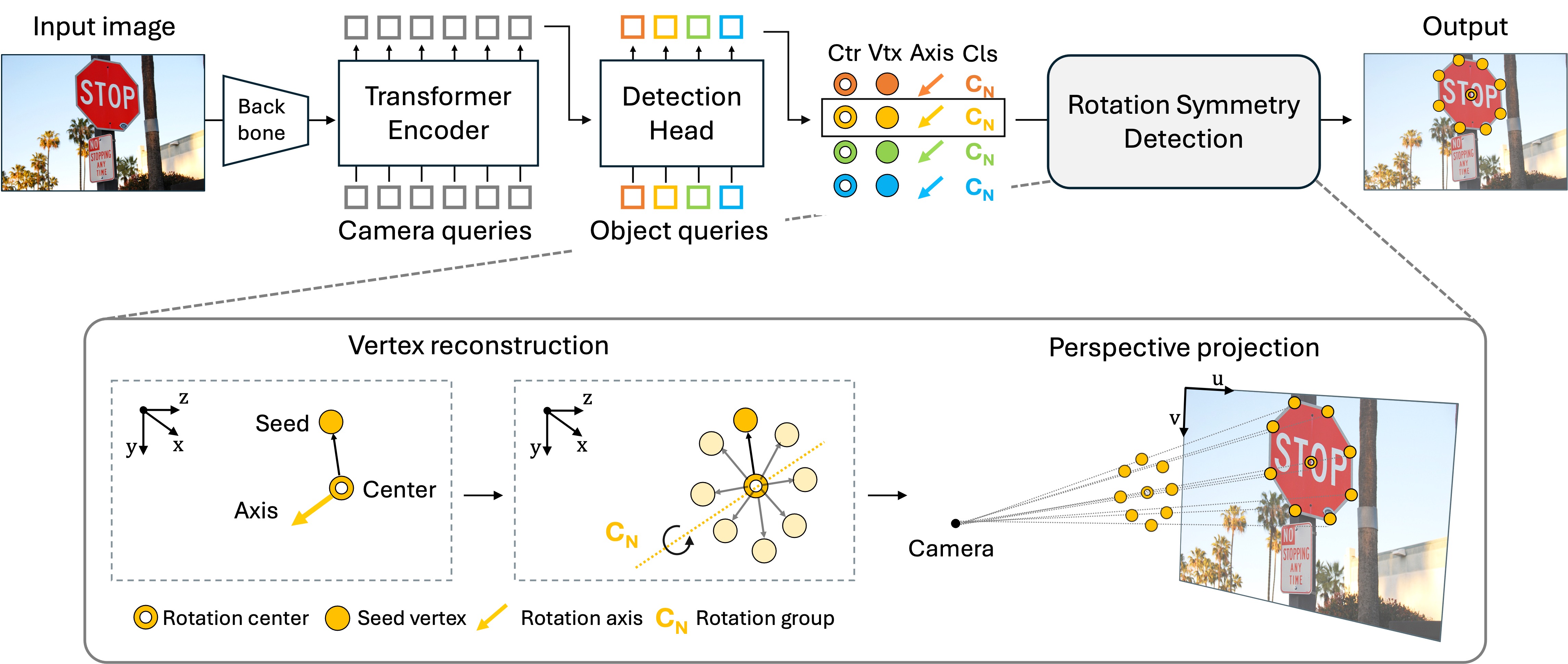}
   \caption{\textbf{Overall pipeline.} 
    The input image is processed through a backbone and transformer encoder with camera queries. The detection head predicts the 3D rotation center, seed vertex, rotation axis, and symmetry group. The seed vertex is then duplicated according to the predicted symmetry group before the 3D coordinates are projected to 2D.
   }
  \label{fig:pipeline}
\vspace{-2mm}
\end{figure*}

Rotation symmetry detection involves predicting the rotation center and supporting vertices, with annotations designed to reflect perceived 3D symmetry rather than just the 2D shape appearance. This paper introduces a novel rotation symmetry detection model that integrates 3D geometric priors to enhance accuracy and robustness.
The model processes the input image using a backbone network followed by a transformer encoder, augmented with additional camera queries. The detection head predicts the 3D rotation center, a seed vertex, the rotation axis vector, and a classification score for the rotation group. The symmetry detection module incorporates vertex reconstruction and perspective projection: the seed vertex is duplicated based on the predicted symmetry group and axis, ensuring evenly spaced vertices around the center while maintaining geometric consistency. Finally, the reconstructed 3D coordinates are projected back to 2D image space. An overview of the complete pipeline is shown in~\Fig{pipeline}.

\subsection{Feature Learning}
\paragraph{Camera queries.}  
The rotation symmetry detector predicts rotation centers and vertices in 3D camera coordinates. To transform backbone features from image coordinates to camera coordinates, we introduce \textit{camera queries}—a set of grid-shaped learnable parameters denoted as \( \mathbf{Q} \in \mathbb{R}^{C \times N_x \times N_y} \). Here, \( N_x \) and \( N_y \) represent the spatial dimensions along the \( x \)- and \( y \)-axes, while \( C \) is the embedding dimension. Each query \( \mathbf{Q}_q \in \mathbb{R}^{C} \), located at \( \mathbf{p}_q \), corresponds to a grid cell in the camera’s local coordinate space, covering a predefined range along the \( x \)- and \( y \)-axes.

\begin{figure}[t]
    \centering
    \includegraphics[width=0.47\textwidth]{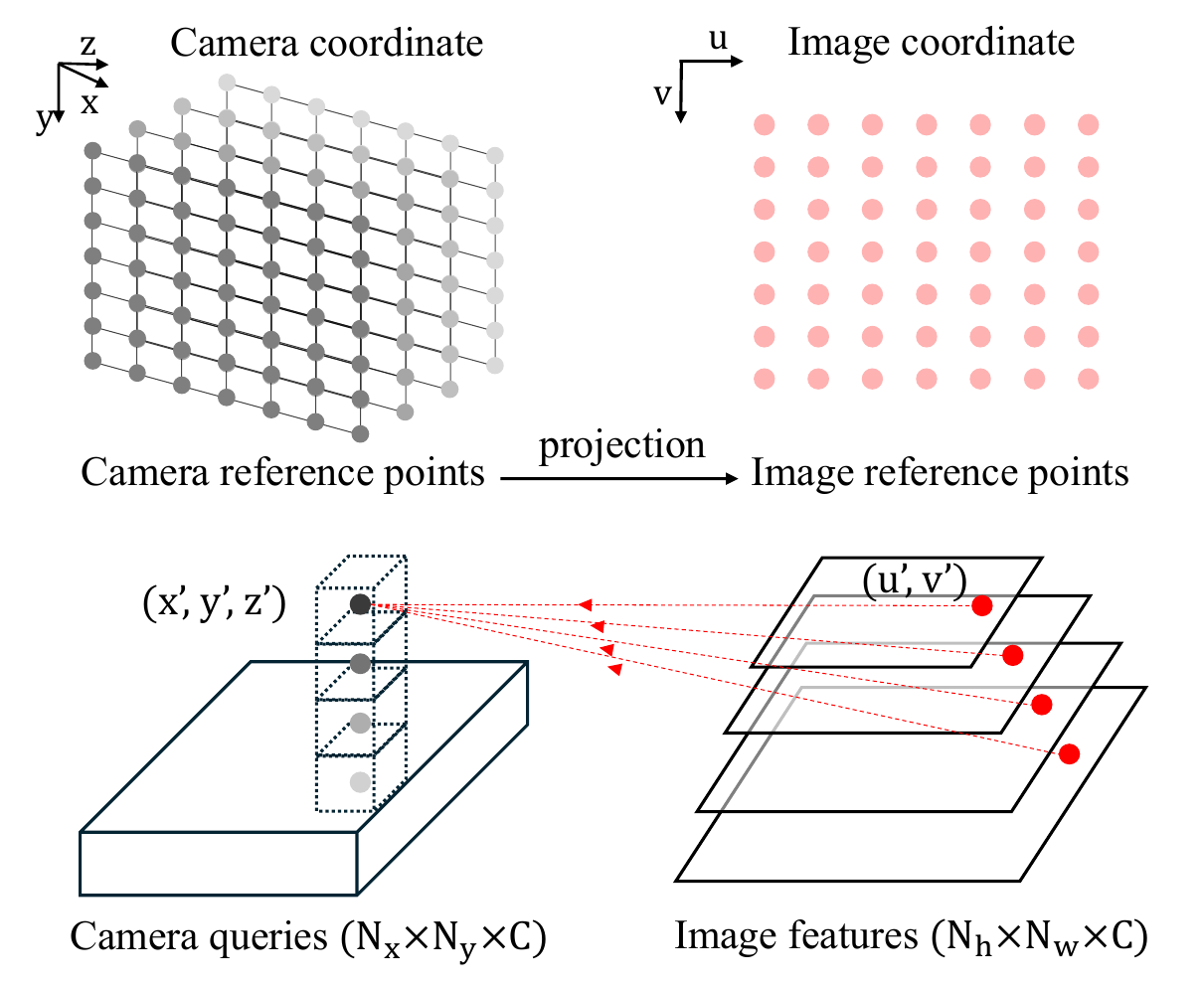}
   \caption{\textbf{Camera Cross Attention.} The 3D reference point grids in camera coordinates are projected onto image coordinates to query the backbone image features.}
\label{fig:attention}
\end{figure}

\paragraph{Camera Cross Attention (CCA).}
Given an input feature map $ \mathbf{F} \in \mathbb{R}^{C \times H \times W} $, let $ q $ index a query feature map $ \mathbf{Q} $ with a 2D reference point $ \mathbf{p}_q $. The camera cross attention is computed as:
\begin{align}
    \text{CCA}(\mathbf{Q}, q, \mathbf{F}) &= 
    \sum^{N_\mathrm{ref}}_{i=1} {\mathrm{Deform}}(\mathbf{Q}_q, \mathcal{P}(\mathbf{p}_q, z_i), \mathbf{F}), \\
    \mathcal{P}(\mathbf{p}_q, z_i) &=  
    \begin{pmatrix} f & 0 & c_x \\ 0 & f & c_y \end{pmatrix} 
    \begin{pmatrix} \sfrac{\mathbf{p}_{q,x}}{z_i} \\ \sfrac{\mathbf{p}_{q,y}}{z_i} \\ 1 \end{pmatrix},
\end{align}
where \( \mathrm{Deform} \) denotes deformable attention~\cite{zhudeformable}, \( f \) is the focal length, and \( c_x, c_y \) is the focal center. For each \( x \)-\( y \) position, \( N_\mathrm{ref} \) depth values along the \( z \)-axis generate 3D reference points, which are projected to 2D to sample image features. See~\Fig{attention} for details.

\paragraph{Transformer encoder.}
The model employs six transformer encoder layers~\cite{zhudeformable}, taking camera queries and multi-scale backbone features as inputs. Multi-scale deformable attention handles self-attention, while the proposed CCA performs cross-attention, linking camera queries with backbone features. Inspired by BEVFormer~\cite{li2022bevformer}, this setup efficiently encodes 2D image features into 3D camera coordinates.
\begin{figure*}[t]
    \centering
    \includegraphics[width=\textwidth]{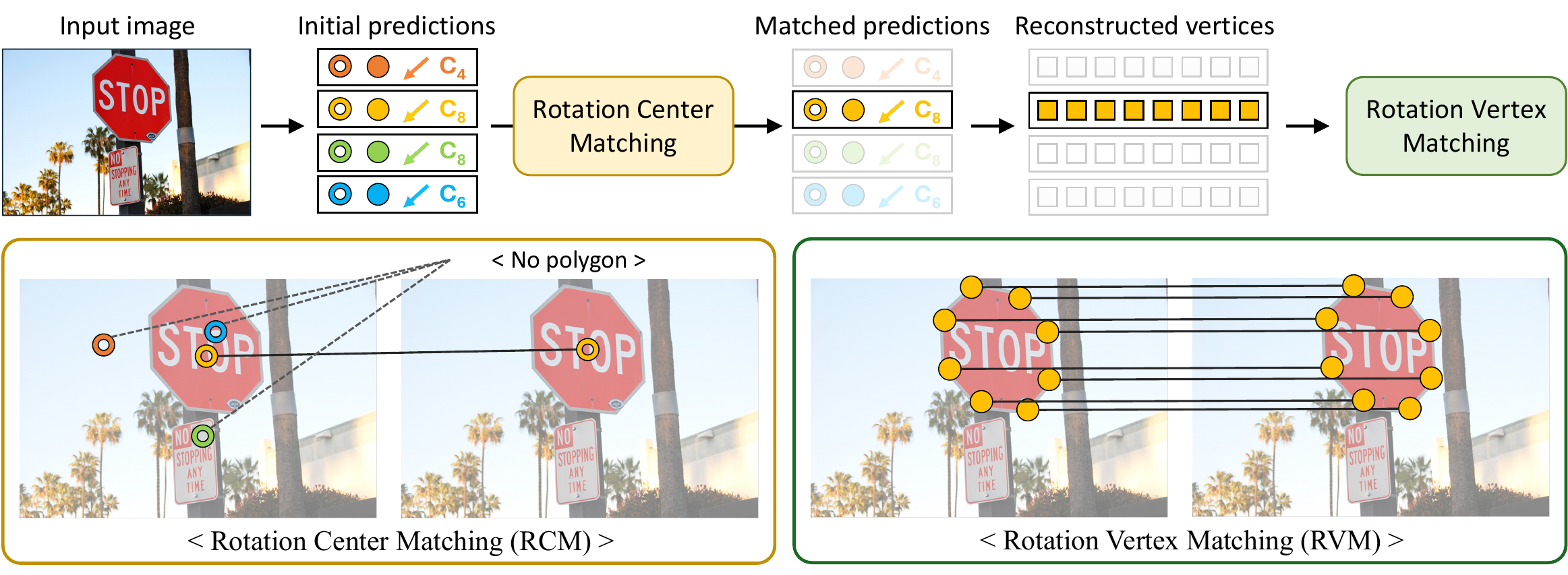}
   \caption{
   \textbf{Overview of the training pipeline.} 
The model employs a two-step bipartite matching process: first, rotation center matching aligns the centers of predictions with ground truth centers, followed by matching the rotation vertices after vertex reconstruction.
}  \label{fig:pipeline_matching}
\vspace{-2mm}
\end{figure*}

\subsection{Rotation Symmetry Detection}
\paragraph{Detection head.}
Each query is processed by the transformer decoder and then passed to a classification branch and a regression branch. The classification branch predicts the rotation symmetry group $g$, defining the order of symmetry $N$. The regression branch outputs four parameters that define the 3D geometric structure as
\begin{align}
\begin{bmatrix} \mathbf{c}^\top \ \mathbf{s}^\top \ \mathbf{a}^\top \ \beta \end{bmatrix}^\top,
\end{align}
where $\mathbf{c} \in \mathbb{R}^3$ is the 3D center coordinate, $\mathbf{s} \in \mathbb{R}^3$ is the 3D seed coordinate (a vertex on the polygon boundary), $\mathbf{a} \in \mathbb{R}^3$ is the 3D axis vector defining the rotation axis, and $\beta$ is an angle bias for initial alignment in certain shapes (e.g., rectangles). These parameters define the spatial structure and orientation needed to construct the polygon in 3D space.

\paragraph{Vertex reconstruction.}
To position vertices according to the predicted rotation symmetry group in 3D space, each vertex \( \mathbf{v}_k \) is computed by rotating a seed point \( \mathbf{s} \) around a rotation axis vector \( \mathbf{a} \), centered at the rotation center \( \mathbf{c} \). The axis \( \mathbf{a} \) is normalized to a unit vector. Each rotation vertex is given by \( \mathbf{v}_k = \mathbf{r}_k + \mathbf{c} \), where the rotated vector \( \mathbf{r}_k \) is calculated using Rodrigues' rotation formula:
\begin{align}
\mathbf{r}_k = \mathbf{r} \cos \theta_k 
+ (\mathbf{a} \times \mathbf{r}) \sin \theta_k 
+ \mathbf{a} (\mathbf{a} \cdot \mathbf{r}) (1 - \cos \theta_k),
\end{align}
with the initial radial vector \( \mathbf{r} = \mathbf{s} - \mathbf{c} \), and rotation angle \( \theta_k = \frac{2\pi k}{N} \). Given the predicted symmetry group order \( N \), we generate \( N \) vertices by setting \( k = 1, 2, \ldots, N \).

\paragraph{Vertex reconstruction for $ \text{C}_2 $ group.}
For the $ \text{C}_2 $ group, vertex reconstruction can be performed by rotating the seed vertex 180 degrees to obtain two vertices. However, $ \text{C}_2 $ objects are typically rectangles consisting of two $ \text{C}_2 $ vertex pairs that share a single center. To efficiently predict these structures within the current framework, we modify the approach to predict four vertices from one $ \text{C}_2 $ center. In this setup, the regression branch additionally predicts an angle bias $ \beta $ (ranging from 0 to 90 degrees). Vertex reconstruction is then performed with rotation angles of $ 0 $, $ \frac{\pi}{4} + \beta $, $ \frac{\pi}{2} $, and $ \frac{\pi}{2} + \beta $ to obtain the vertices from the seed vertex.

\paragraph{3D-to-2D projection.}
The predicted center $\mathbf{c}$ and vertices $\mathbf{v}_1, \cdots, \mathbf{v}_
V$ are projected onto the 2D image plane using the focal length $f$ and focal center $(c_x, c_y)$. The projection of a 3D point $\mathbf{p} = (p_x, p_y, p_z)^\top$ to its 2D image coordinates $\mathbf{p}' = (p'_x, p'_y)^\top$ is expressed in matrix-vector form as
\begin{align}
\begin{pmatrix} p'_x \\ p'_y \end{pmatrix} = 
\begin{pmatrix} f & 0 & c_x \\ 0 & f & c_y \end{pmatrix} 
\begin{pmatrix} \sfrac{p_x}{p_z} \\ \sfrac{p_y}{p_z} \\ 1 \end{pmatrix}.
\end{align}
By applying this projection to the center $\mathbf{c}$ and each vertex $\mathbf{v}_k$, the 2D image coordinates of the rotation center $\mathbf{c}'$ and vertices $\{\mathbf{v}'_k\}_{k=1}^{V}$ are obtained.

\subsection{Training}
\label{training}

\paragraph{Set-based training.}
Each image contains a ground-truth set of polygons $\mathcal{Y}$. Each polygon $\mathcal{Y}_i=(\mathbf{c}'_i, g_i, \mathcal{V}_i)$ 
where a rotation center coordinate $\mathbf{c}'$, a rotation group $g$, and 
a set of vertices $\mathcal{V} = \{\mathbf{v}'_k\}_{k=1}^{V}$, where $V$ is the number of vertices defined by the rotation group. The set of predictions, denoted as $\hat{\mathcal{Y}}$, each prediction additionally includes a binary classification logit for each group $g$ as $\hat{p}(g)$. 
During training, a two-step matching process optimizes the detection of rotation centers and vertices. First, the model aligns predicted polygon centers with ground-truth centers. For each matched polygon, the model then aligns the set of vertices. 

\paragraph{Rotation Center Matching (RCM).}
The optimal assignment $ \hat{\sigma} $ is computed to minimize the matching cost $\mathcal{C}_{\text{center}}$:
\begin{align}
\hat{\sigma} &= \arg\min_{\sigma} \sum\nolimits_{i} \mathcal{C}_{\text{center}}(\mathcal{Y}_i, \hat{\mathcal{Y}}_{\sigma(i)}), 
\end{align}
where the matching cost is defined as
\begin{align}
\mathcal{C}_{\text{center}} &= \mathds{1}_{\{g_i \neq \emptyset\}} \big[- \hat{p}_{\sigma(i)}(g_i) + \|\mathbf{c}'_i - \hat{\mathbf{c}}'_{\sigma(i)}\|_1 \big].
\end{align}
Following~\cite{carion2020detr}, the ground-truth set $\mathcal{Y}$ is augmented to include $N$ polygons by inserting $\emptyset$ for missing entries.

\paragraph{Rotation Vertex Matching (RVM).}
Once the centers are matched, the corresponding vertex sets $(\mathcal{V}i, \hat{\mathcal{V}}{\sigma(i)})$ are aligned. The optimal vertex assignment $ \hat{\rho} $ is defined to minimize the vertex matching cost as
\begin{align}
\hat{\rho} &= \arg\min_{\rho} \sum_{j=1}^{V} \| \mathbf{v}'_j - \hat{\mathbf{v}}'_{\rho(j)}\|_1.
\end{align}
Only predictions matched with $\mathcal{Y}_i$ with non-empty $g_i$ are included in the vertex matching process.

\paragraph{Training Objective.}
For each polygon pair $(y, \hat{y})$, the center loss $ \mathcal{L}_{\text{center}} $ and the vertex loss $ \mathcal{L}_{\text{vertex}} $ are defined as
\begin{align}
\mathcal{L}_{\text{center}}(\mathcal{Y}_i, \hat{\mathcal{Y}}_{\sigma(i)}) &= - \log \hat{p}_{\sigma(i)}(g_i) + \mathcal{L}_{\text{reg}}(\mathbf{c}'_i, \hat{\mathbf{c}}'_{\sigma(i)}), \\
\mathcal{L}_{\text{vertex}}(\mathcal{V}_i, \hat{\mathcal{V}}_{\sigma(i)}) &= \sum\nolimits_{k} \mathcal{L}_{\text{reg}} (\mathbf{v}'_{k}, \hat{\mathbf{v}}'_{\hat{\rho}(k)}),
\end{align}
where $\mathcal{L}_{\text{reg}}$ indicates the $\textit{l}_1$ loss.
The overall training objective $\mathcal{L}_{\text{total}}$ is defined as
\begin{align}
\mathcal{L}_{\text{total}} &= 
\sum\nolimits_i \mathcal{L}_{\text{center}}(\mathcal{Y}_i, \hat{\mathcal{Y}}_{\sigma(i)}) +  
\mathcal{L}_{\text{vertex}}(\mathcal{V}_i, \hat{\mathcal{V}}_{\sigma(i)}).
\end{align}

\section{Experiments}
\label{experiment}
\begin{figure}[t]
    \centering
    \includegraphics[width=0.47\textwidth]{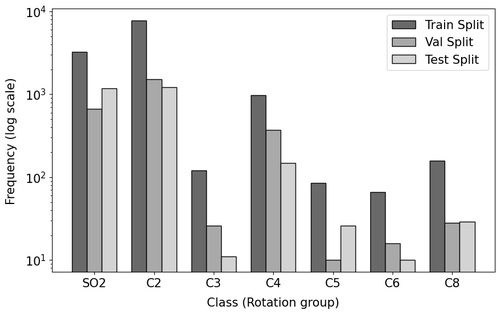}
   \caption{
   Class distribution of rotational symmetry groups in the DENDI dataset, shown as object counts (log scale) per group. The x-axis shows rotation groups, and the y-axis shows object frequency. Train, validation, and test splits are shown side-by-side.
}
\label{fig:class_distribution}
\vspace{-2mm}
\end{figure}

\subsection{Experimental Settings}
\paragraph{Datasets.}
The DENDI dataset~\cite{seo2022equisym} is a real-world dataset designed for symmetry detection. The rotational symmetry subset consists of 1,459 images for training, 313 for validation, and 307 for testing. It includes annotated rotation centers for objects belonging to the following rotational symmetry groups: SO(2), $ \text{C}_2 $, $ \text{C}_3 $, $ \text{C}_4 $, $ \text{C}_5 $, $ \text{C}_6 $, and $ \text{C}_8 $. For polygonal objects with cyclic rotational symmetry, vertex coordinates are also provided. Only discrete rotation groups are used for vertex detection, resulting in 219 validation images and 253 testing images. The class distribution across symmetry groups is highly imbalanced, as shown in~\Fig{class_distribution}.

\paragraph{Evaluation.}
Average Precision (AP) is used to evaluate rotation symmetry center and vertex detection. A prediction is considered correct if \( d < \tau \cdot \max(w, h) \), where \( d \) is the distance to the nearest ground-truth center, \( (w, h) \) are the image dimensions, and \( \tau = 0.025 \)~\cite{liu2013symmetry}. During evaluation, ground-truth and predicted centers are matched using bipartite matching, followed by vertex matching for each pair. Vertex AP is computed only for successfully matched polygons.  
To compare with prior work, we also report the F1-score of predicted rotation centers. Since this metric uses score maps, predicted centers are rendered as heatmaps with confidence scores. Following~\cite{seo2022equisym}, ground-truth and predicted score maps are dilated by 5 pixels.

\paragraph{Implementation details.}
We use a Swin-T backbone~\cite{Liu_2021_swin} pretrained on ImageNet~\cite{deng2009imagenet}, mapping features from layers 1–3 to 256 dimensions. The model is trained for 200 epochs using AdamW~\cite{loshchilov2018decoupledAdamW} with a learning rate of 0.0002, weight decay of 0.001, and a backbone learning rate 10× lower. Training uses a batch size of 8 with images resized to a max side of 1333 pixels, and only random flipping is applied for rotation augmentation.
The model predicts 3D points within a point cloud range of \([-1, 1]\) meters in x-y and \([0, 4]\) meters in z. Since the DENDI dataset~\cite{seo2022equisym} lacks camera intrinsics, we set the focal center to the image center and the focal length to 1000. The regression cost is weighted at 10, while center and vertex losses are weighted at 1. We use 800 object queries and apply no test-time post-processing.

\subsection{Evaluation of the Proposed Method}
\begin{table}[t]
\centering
\caption{Rotation center detection results on DENDI.}
\label{tab:center-eval}
\vspace{-2mm}
\setlength{\tabcolsep}{0.3em}
\begin{tabular}{lcccccccc}
\toprule
Method & SO(2) & $\text{C}_2$ & $\text{C}_3$ & $\text{C}_4$ & $\text{C}_5$ & $\text{C}_6$ & $\text{C}_8$ & mean\\
\midrule
Ours & 62.1 & 30.7 & 6.3 & 5.3 & 48.6 & 9.0 & 65.9 & 31.3 \\
\bottomrule
\vspace{-2mm}
\end{tabular}
\end{table}

\begin{table}[t]
\centering
\caption{Rotation vertex detection results on DENDI.}
\label{tab:vertex-eval}
\vspace{-2mm}
\setlength{\tabcolsep}{0.3em}
\begin{tabular}{lccccccc}
\toprule
Method & $\text{C}_2$ & $\text{C}_3$ & $\text{C}_4$ & $\text{C}_5$ & $\text{C}_6$ & $\text{C}_8$ & mean\\
\midrule
Ours & 39.4 & 14.3 & 27.1 & 27.3 & 28.8 & 46.7 & 30.6 \\
\bottomrule
\vspace{-2mm}
\end{tabular}
\end{table}

\paragraph{Rotation center detection.}
Rotation center detection results on the DENDI test set are shown in~\Tbl{center-eval}. Performance for the \( \text{C}_3 \) and \( \text{C}_6 \) groups is relatively low, likely due to limited testing samples. For \( \text{C}_4 \), performance is affected by potential confusion with \( \text{C}_2 \) and the smaller size of the objects, which can lower AP.

\paragraph{Rotation vertex detection.}
Rotation vertex detection results on the DENDI test set are presented in~\Tbl{vertex-eval}. The SO(2) group is excluded from both training and evaluation. For \( \text{C}_N \) groups (except \( \text{C}_2 \)), objects are annotated with \( N \) vertices. For \( \text{C}_2 \), only rectangular objects are included, excluding ellipses. Each image uses 800 object queries to predict polygons. Predicted polygons are first matched to ground-truth polygons based on center proximity. Vertex-level true positives are then determined by bipartite matching between the vertex sets of each matched pair, followed by classification using a distance threshold.

\begin{figure*}[t]
    \centering
    \includegraphics[width=\textwidth]{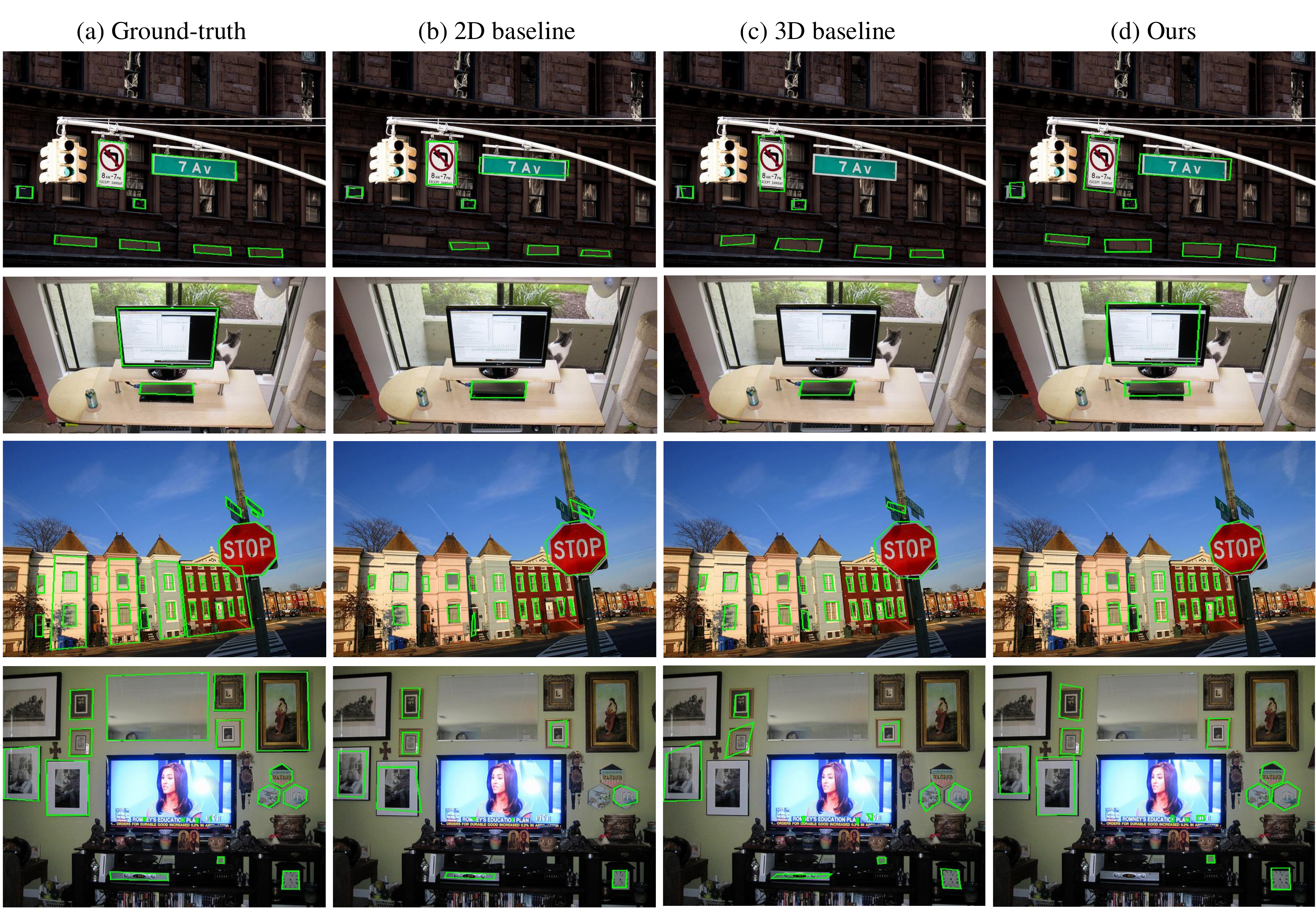}
   \caption{
   \textbf{Qualitative comparison of rotation vertex detection results on the DENDI dataset.}
      Each set of four columns displays ground truth, the results of 2D baseline, 3D baseline, and ours. Only polygon predictions with all true-positive vertices are marked(green).
}  \label{fig:qual_vertex}
\vspace{-2mm}
\end{figure*}

\subsection{Ablation Study}
\paragraph{Baseline models.}
To evaluate the impact of each component on rotation vertex detection, we compare three models: a 2D baseline, a 3D baseline, and our full model. The 2D baseline directly predicts rotation centers and vertices in image coordinates. The 3D baseline introduces camera cross attention to lift 2D features into 3D space, allowing predictions in 3D coordinates that are then projected back to 2D. Our final model includes a vertex reconstruction stage, where a seed vertex is duplicated around the center based on the predicted symmetry group for geometric consistency.

\begin{table}[t]
\centering
\caption{Rotation vertex detection results on DENDI.}
\label{tab:ablation}
\vspace{-2mm}
\setlength{\tabcolsep}{0.3em}
\begin{tabular}{lccc}
\toprule
Method & 3D query/pred. & vertex recon.  & mAP \\
\midrule
2D baseline & &  & 24.7 \\
3D baseline & \checkmark &  & 23.5 \\
\midrule
Ours & \checkmark & \checkmark  & \textbf{30.6} \\
\bottomrule
\vspace{-3mm}
\end{tabular}
\end{table}

\paragraph{Quantitative results.}
As shown in~\Tbl{ablation}, the 2D baseline achieves a mean AP of 24.7, while the naive 3D baseline drops slightly to 23.5, likely due to projection noise or misalignment without strong geometric constraints. With our vertex reconstruction scheme, which enforces symmetry-aware 3D priors, performance significantly improves to 30.6. 
These results demonstrate the importance of explicitly modeling 3D geometric constraints to enhance both accuracy and structural consistency in rotation vertex detection.

\paragraph{Qualitative results.}
\Fig{qual_vertex} shows a qualitative comparison of the baselines and our method. Our model follows geometric rules like equal side lengths and angles by reconstructing vertices instead of predicting them individually. 
As seen in the last row, our method handles cluttered scenes well and performs notably better on \( \text{C}_6 \) objects.

\begin{figure*}[t]
    \centering
    \includegraphics[width=\textwidth]{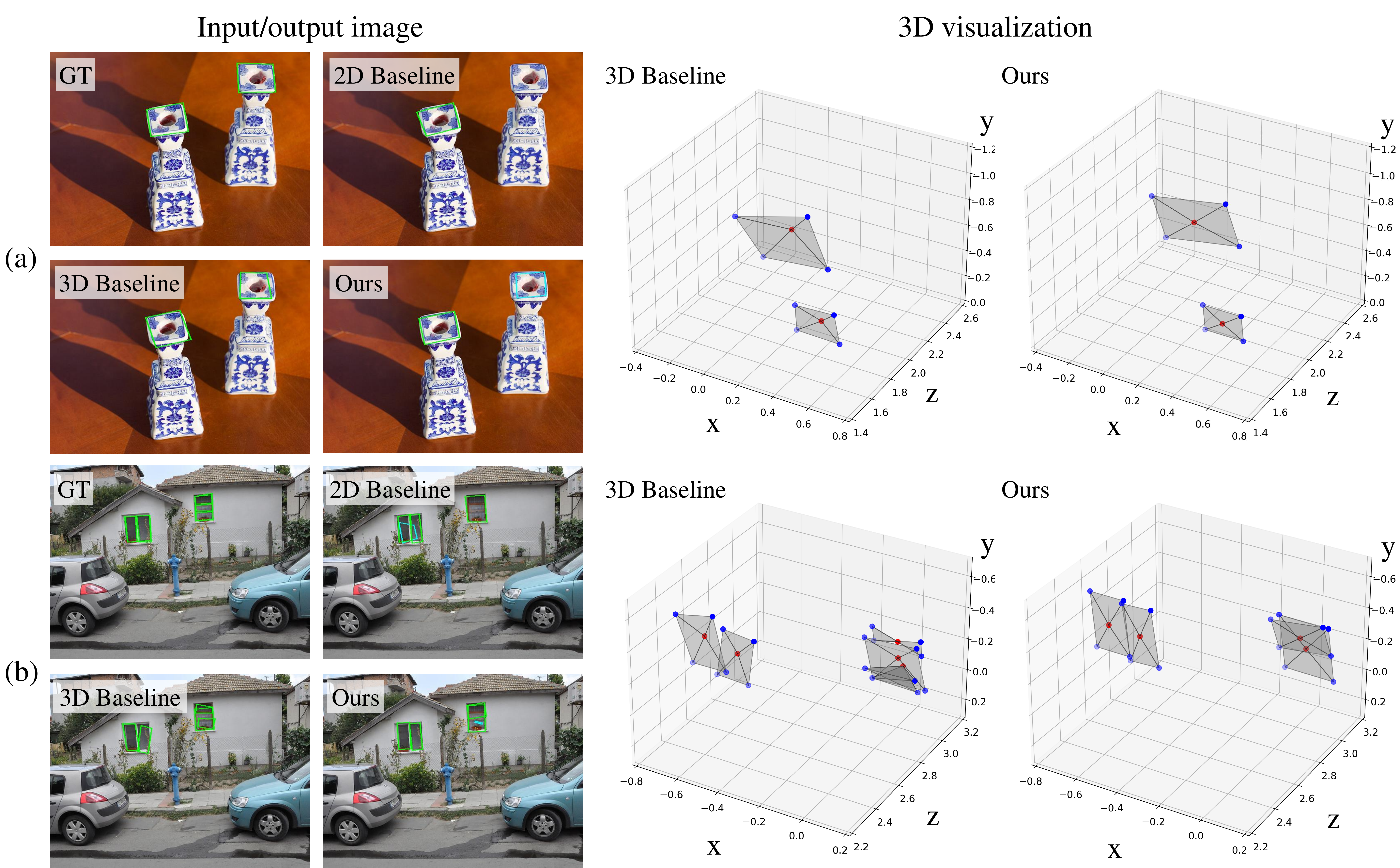}
    \vspace{1mm}
   \caption{
   \textbf{Qualitative comparison of rotation vertex detection on the DENDI dataset.}
   Each row shows input images with detected vertices (true positives in green, others in cyan) for the ground truth, 2D baseline, 3D baseline, and our method, followed by 3D plots illustrating predicted points and geometric properties such as distances and co-planarity.
}  \label{fig:qual_3dplot}
\end{figure*}

\subsection{Analysis in 3D}  
To evaluate the effect of 3D geometric priors, we compare our model against a 3D baseline that predicts 3D points and projects them into 2D in~\Fig{qual_3dplot}. Each row includes the input, ground truth, 2D baseline, 3D baseline, and our result, followed by 3D plots of predicted points. Detected vertices are overlaid on input images (green for true positives, cyan for others), while 3D plots show the rotation center (red), vertices (blue), and a triangle connecting the center to two nearby vertices to illustrate geometric alignment.

In row (a), the object appears tilted. While the 3D baseline provides slightly better 2D localization, its 3D predictions show distorted vertex placements. Our model, by contrast, correctly captures the rectangular structure in both 2D and 3D, benefiting from geometric priors. In row (b), our model accurately detects overlapping rectangular windows, particularly those closer to the camera, preserving consistent 3D spacing and depth ordering—something the baseline struggles with.
These examples demonstrate that while 2D accuracy alone does not guarantee proper 3D arrangement, incorporating 3D geometric constraints improves both spatial structure and projection quality. Nonetheless, some failure cases remain, particularly in estimating the rotation axis under severe viewpoint variations.

\begin{figure*}[t]
    \centering
    \includegraphics[width=\textwidth]{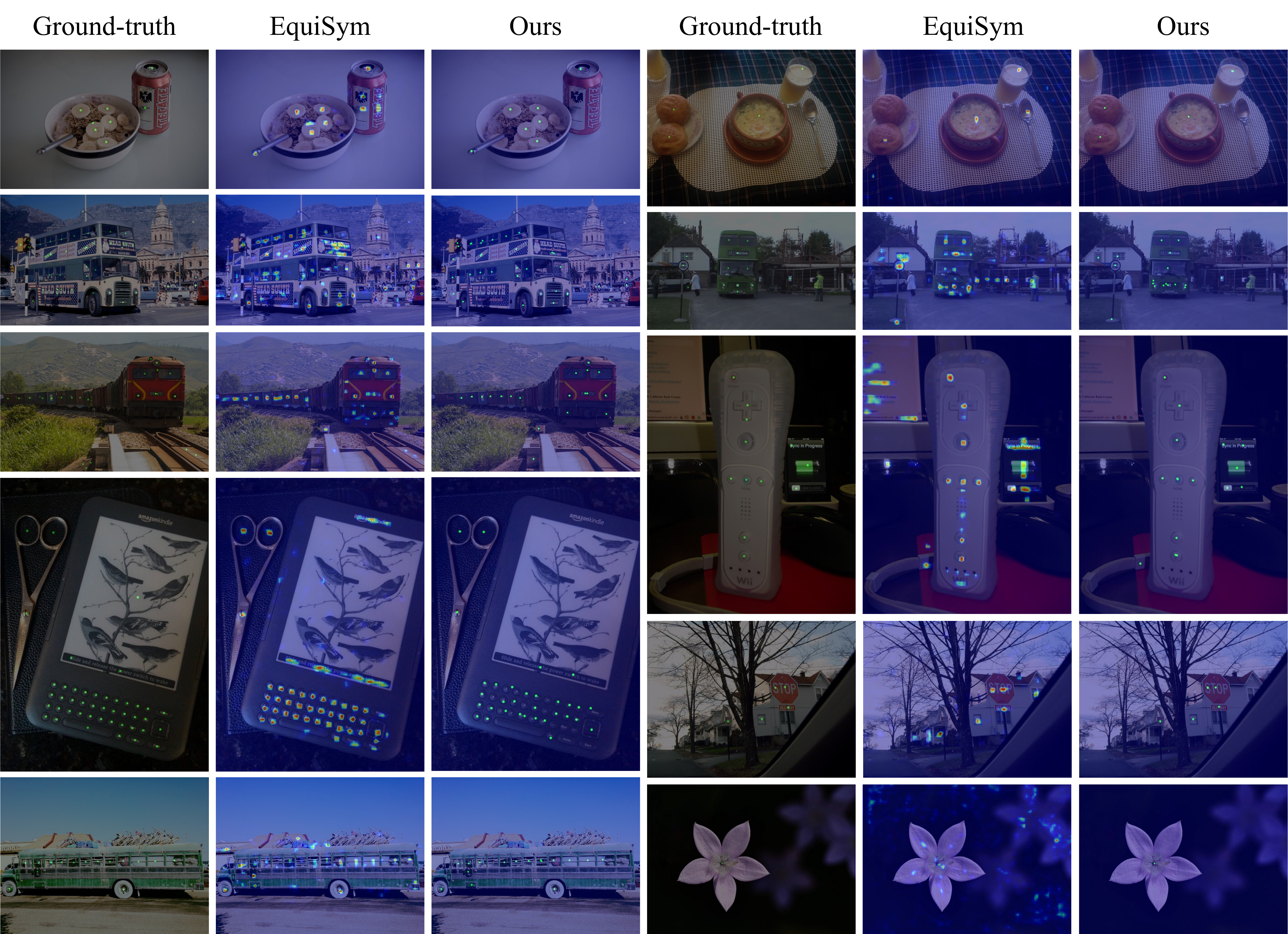}
   \caption{
   \textbf{Qualitative results of rotation center detection on the DENDI dataset.}
   Each set of three columns shows ground truth, EquiSym~\cite{seo2022equisym}, and our method. Our detection-based model allows for analysis of individual symmetries. 
}  \label{fig:qual_center}
\vspace{-2mm}
\end{figure*}

\begin{table}[t]
\centering
\caption{Rotation symmetry detection results on DENDI.}
\label{tab:sota-eval}
\vspace{-2mm}
\setlength{\tabcolsep}{0.3em}
\begin{tabular}{lccccc}
    \toprule
    {Method}   &{prediction}      & {max f1-score} \\
    \midrule
    $\text{EquiSym}$~\cite{seo2022equisym} & segmentation  & 22.5       \\
    Ours & detection   & \textbf{33.2}\\
    \bottomrule
    \end{tabular}
\end{table}

\subsection{Comparison with the State-of-the-art Method}
\paragraph{Quantitative results.}
We compare our approach with the state-of-the-art method by evaluating it within a heat-map based framework, as presented in~\Tbl{sota-eval}. Our model outperforms EquiSym~\cite{seo2022equisym} with a maximum F1-score of 33.2. To produce the foreground score map, predicted points for all classes that exceed the score threshold are rendered. Consistent with~\cite{seo2022equisym}, 100 thresholds between 0 and 1 are applied. Remarkably, despite employing a detection-based approach, our model surpasses EquiSym’s segmentation metric, highlighting the robustness of the proposed method even when evaluated with segmentation-based metrics.

\paragraph{Qualitative results.}
Detected rotation centers are shown in~\Fig{qual_center}, using a confidence threshold that maximizes the F1-score in~\Tbl{sota-eval}. Unlike EquiSym, which outputs heatmaps requiring post-processing, our detection-based approach directly identifies individual symmetries. This not only aligns with traditional methods but also enables various downstream applications. The results also show the model’s ability to detect dense symmetries across groups.

\section{Conclusion}
\label{conclusion}
We have presented a rotation symmetry detection model that leverages 3D geometric priors for robust symmetry analysis.
It directly predicts rotation centers and vertices in 3D space and projects the results back to 2D while preserving structural integrity. Experiments demonstrate the effectiveness of 3D geometric priors in the detection of rotation vertices.
Future work will address challenging cases with viewpoint variations by predicting camera intrinsics to improve projection accuracy. \\
\vspace{-2mm}
\smallbreak
\noindent \textbf{Acknowledgements.}
This work was supported by the Samsung Electronics AI Center and also by the IITP grants (RS-2022-II220290: Visual Intelligence for Space-Time Understanding and Generation (50\%), RS-2021-II212068: AI Innovation Hub (45\%), RS-2019-II191906: Artificial Intelligence Graduate School Program at POSTECH (5\%)) funded by the Korea government (MSIT).
\looseness=-1

{
    \small
    \bibliographystyle{ieeenat_fullname}
    \bibliography{main}
}

\clearpage
\appendix

\setcounter{section}{0}
\setcounter{figure}{0}
\setcounter{table}{0}

\renewcommand{\thesection}{A.\arabic{section}}
\renewcommand{\thefigure}{a.\arabic{figure}}
\renewcommand{\thetable}{a.\arabic{table}}
\section*{Appendix}

\section{Use of Estimated Camera Intrinsics}

\begin{figure}[h]
    \centering
    \includegraphics[width=0.45\textwidth]{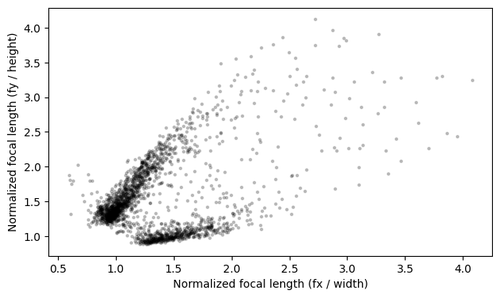}
\vspace{-2mm}
   \caption{
   Normalized focal length distribution of DENDI.
}  
\label{fig:focal_plot}
\end{figure}

In the main paper, we used a fixed focal length of 1000 due to missing camera intrinsics. To improve this, we apply DiffCalib\footnote{He, Xiankang, et al. "DiffCalib: Reformulating Monocular Camera Calibration as Diffusion-Based Dense Incident Map Generation." \textit{arXiv preprint}, 2024.}, a diffusion-based model that estimates intrinsics from monocular images. Using DiffCalib, we estimate focal lengths for each DENDI image. These values vary widely, as shown in~\Fig{focal_plot}, so we clip the normalized focal lengths to the range \([2/3, 4/3]\) for training stability.
Our original 3D model used a fixed point cloud range of \([-1, 1]\) m (x-y) and \([0, 4]\) m (z). To accommodate the variation in focal lengths, we expand this range by a factor of 4.

Results in~\Tbl{intrinsic} show that our model achieves 20.5 mAP with 800 scores (same as the main paper), and improves to 25.0 mAP with 3200 scores—the highest among all settings. Although this is lower than the 30.6 mAP in Table 4, the drop is due to unstable intrinsics and unoptimized hyperparameters. Still, our vertex reconstruction consistently outperforms the 3D baseline in all cases.

\begin{table}[h]
\centering
\caption{Rotation vertex detection results on DENDI \textit{test}.}
\label{tab:intrinsic}
\vspace{-2mm}
\setlength{\tabcolsep}{0.3em}
\begin{tabular}{lcccc}
\toprule
Method & 3D model & 3D prior & mAP$_{800}$ & mAP$_{3200}$ \\
\midrule
2D Baseline & & & 22.6 & 23.1 \\
3D Baseline & \checkmark & & 14.8 & 15.8 \\
\midrule
Ours & \checkmark & \checkmark & 20.5 & 25.0 \\
\bottomrule
\end{tabular}
\end{table}

\section{Analysis in 3D}

\begin{figure*}[h!]
    \centering
    \includegraphics[width=\textwidth]{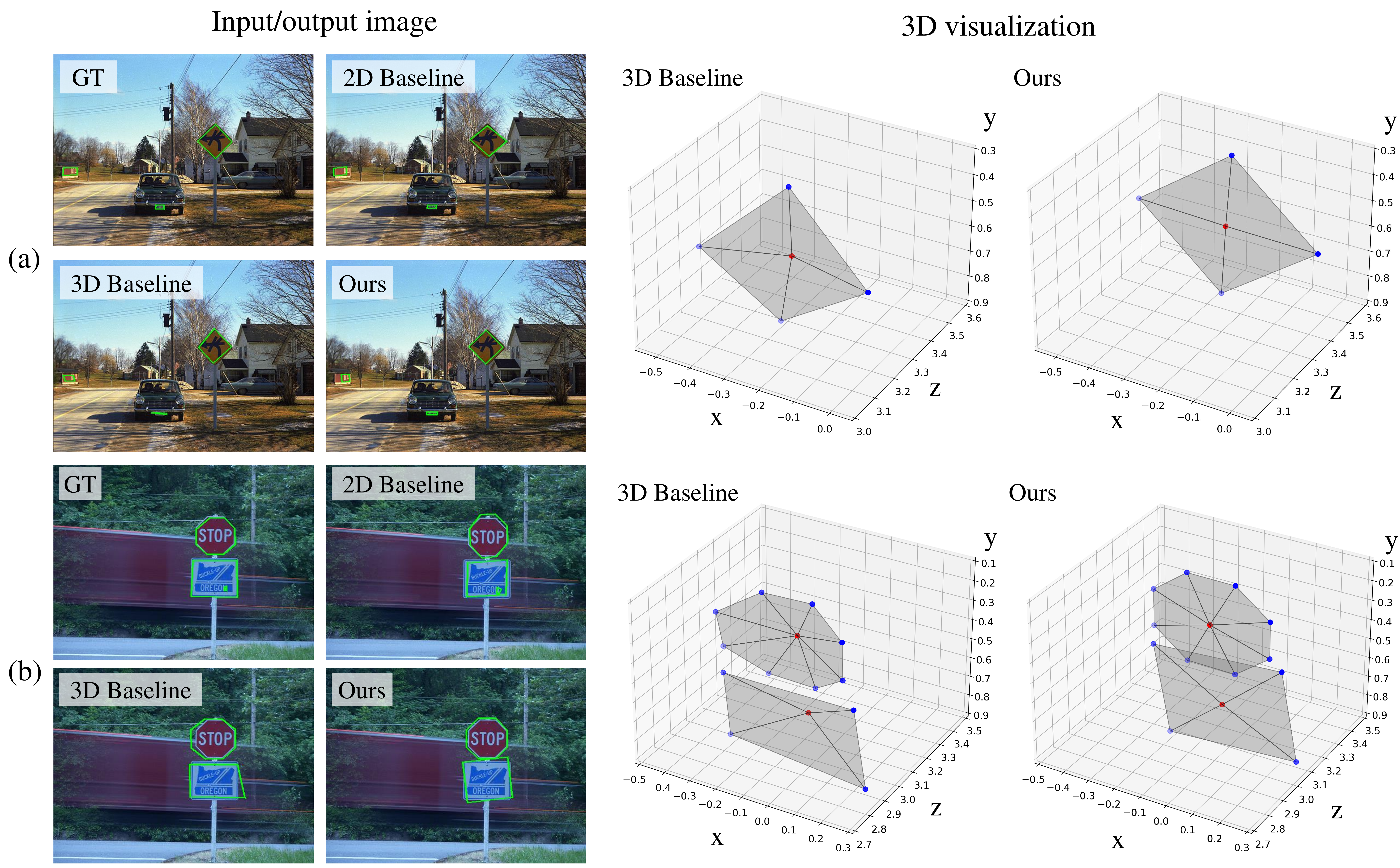}
    \vspace{1mm}
   \caption{
   \textbf{Qualitative comparison of rotation vertex detection results on the DENDI dataset.}
   Each row displays input images with detected vertices (true positives in green, others in cyan) for the ground truth, 2D baseline, 3D baseline, and our method, alongside 3D plots visualizing the predicted points and geometric properties such as distances and co-planarity.
}  \label{fig:qual_plot_supp}
\end{figure*}

To analyze the effectiveness of incorporating explicit 3D geometric priors for rotation symmetry detection, we established a 3D baseline model that predicts 3D points and projects them into 2D image space. The qualitative results in~\Fig{qual_plot_supp} showcase four samples from the DENDI validation and test sets, arranged in rows, each containing six items. In each row, four images (input image, ground truth, 2D baseline, 3D baseline, and ours) are displayed in a 2x2 format, with two plots on the right. 

The accompanying plots visualize the 3D predicted points from the 3D baseline and our method in XYZ camera coordinates. Key points include the rotation center (red dot) and rotation vertices (blue dots). Additionally, triangles formed by the center and two nearby vertices are illustrated, providing insight into the 3D placement of the points, including distances and co-planarity, highlighting the geometric accuracy of our approach.

In row (a), showing a \( \text{C}_4 \) object, all models detect the correct vertices, but ours achieves the best localization. The 3D baseline appears accurate in 2D but fails to maintain correct 3D geometry, highlighting the benefit of our geometric priors. 
In row (b), with \( \text{C}_8 \) and \( \text{C}_2 \) signs, our method performs best on the octagon but lags slightly on \( \text{C}_2 \) compared to the 2D baseline—likely due to the simpler nature of bounding box detection and our still-developing \( \text{C}_2 \) reconstruction.
Overall, our method improves 2D accuracy by enforcing 3D geometric constraints.

\section{Additional Qualitative Results}
\begin{figure*}[t]
    \centering
    \includegraphics[width=\textwidth]{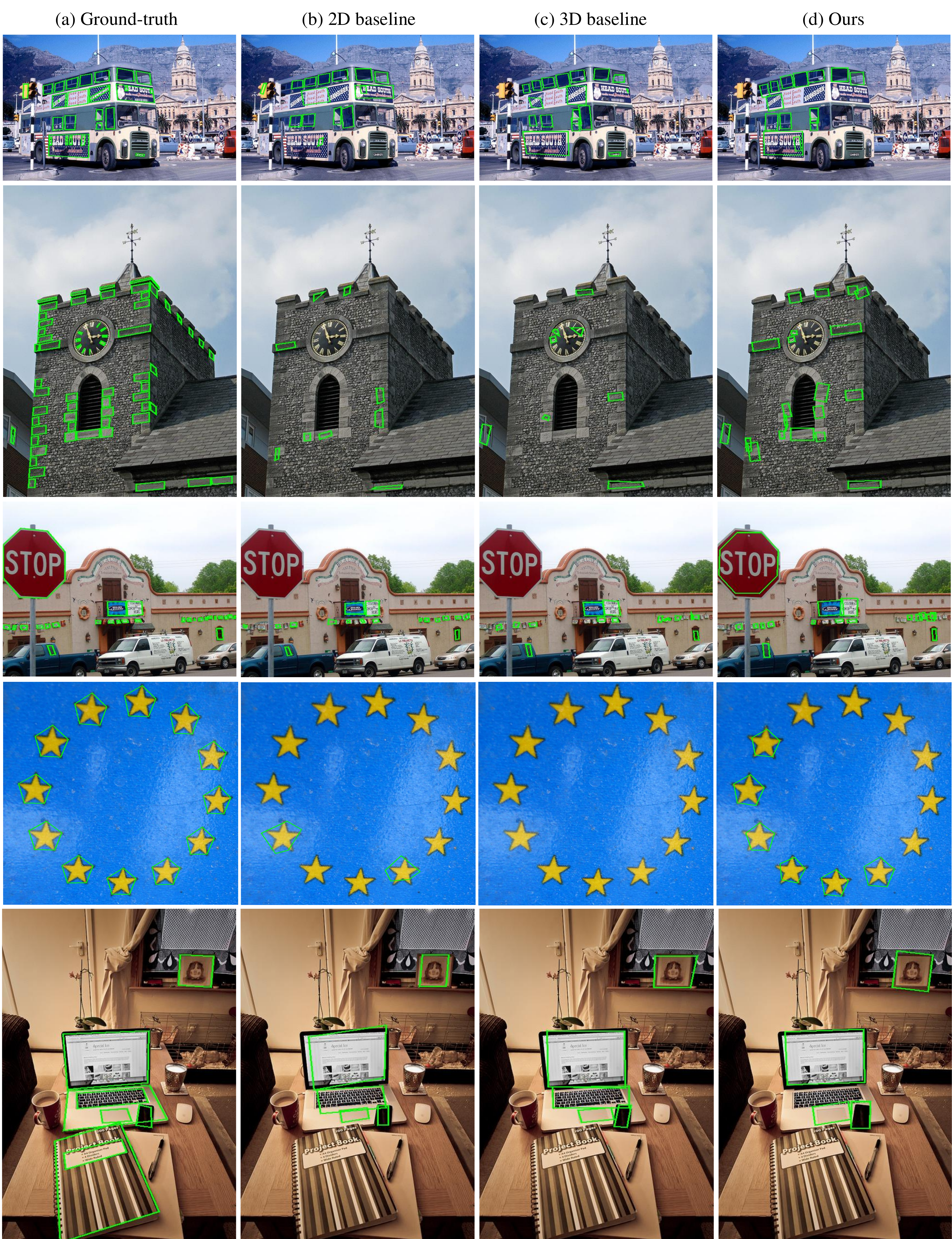}
    \vspace{1mm}
   \caption{
   \textbf{Qualitative comparison of rotation vertex detection results on the DENDI dataset.} 
Each set of four columns displays the ground truth, the 2D baseline, the 3D baseline, and ours. True-positive polygons are marked in green; otherwise, they are marked in cyan.
}  \label{fig:qual_vertex_supp}
\end{figure*}

\Fig{qual_vertex_supp} shows additional qualitative comparisons between the 2D baseline, 3D baseline, and our method. The results demonstrate that our model effectively detects dense rotational symmetries in real-world scenes. By incorporating 3D geometric priors, our method improves localization in the 3D detection setting. Notably, in the fourth row, it outperforms both baselines on objects with \( \text{C}_5 \) symmetry.

\end{document}